%% file: root.tex
\documentclass[letterpaper, 10 pt, conference]{ieeeconf}  

\IEEEoverridecommandlockouts                              

\usepackage{graphicx}
\usepackage[utf8]{inputenc}
\usepackage{graphics,graphicx}
\usepackage{amsfonts}
\usepackage{amsmath}
\usepackage{amsthm}
\usepackage{amssymb}
\usepackage{tikz}
\usepackage{cite}
\usepackage{booktabs}

\allowdisplaybreaks
\theoremstyle{definition}

\usepackage[hidelinks]{hyperref}
\usepackage[noabbrev]{cleveref}
\crefname{equation}{}{} 
\crefname{section}{Sec.}{Sec.}
\crefname{figure}{Fig.}{Figure.}
\usepackage[caption = false]{subfig}

\usepackage{color}
\usepackage{xcolor}

\usepackage[normalem]{ulem}
\usepackage[linesnumbered,ruled,vlined]{algorithm2e}
\SetKwInput{KwInput}{Input}     
\SetKwInput{KwOutput}{Output}  
\usepackage{algpseudocode}
\algnewcommand{\algorithmicand}{\textbf{ and }}
\algnewcommand{\algorithmicor}{\textbf{ or }}
\algnewcommand{\OR}{\algorithmicor}
\useunder{\uline}{\ul}{}

\usetikzlibrary{positioning, arrows.meta}
\tikzset{%
  >={Latex[width=2mm,length=2mm]},
  state/.style={draw=none, circle, inner sep=0pt, minimum size=10mm},
  context/.style={draw, circle, dashed, minimum size=10mm},
  arrow/.style={->, line width=1mm},
  observation/.style={draw, rectangle, minimum size=8mm},
  gp/.style={draw, rectangle, dashed, minimum size=8mm},
}

\title{
\LARGE \bf 
Out of Distribution Detection via Domain-Informed \\Gaussian Process State Space Models
}
\author{Alonso Marco, Elias Morley and Claire J. Tomlin
\thanks{
All the authors are with the Department of Electrical Engineering and Computer Sciences, University of California, Berkeley, 2594 Hearst Ave, Berkeley, CA 94720, USA. {\tt\small amarco@berkeley.edu, emorley@berkeley.edu, tomlin@eecs.berkeley.edu.}
}
\thanks{This work is supported by the DARPA Assured Autonomy program, the NASA ULI Project in Safe Aviation Autonomy, and the Rafael del Pino Foundation.}}

\input{./defs.tex}

\input{./def_tablesfigs.tex}

\begin{document}




\maketitle
\thispagestyle{empty}
\pagestyle{empty}



\input{sec_abstract.tex}

\input{sec_intro.tex}
\input{sec_prelim.tex}
\input{sec_kernelconstruction.tex}
\input{sec_problem.tex}
\input{sec_results.tex}

\input{sec_conclusions.tex}


\bibliography{root}
\bibliographystyle{ieeetr} 

\end{document}

%% file: defs.tex
\newcommand{\zt}{z_t}
\newcommand{\ztp}{z_t^\prime}
\newcommand{\fnom}{f_\textrm{nom}}

\newcommand{\R}{\mathbb{R}}

\newtheorem{example}{Example}



%% file: def_tablesfigs.tex
\newcommand{\graphmodel}[1]{
\begin{figure}[#1]
\centering
\begin{tikzpicture}[node distance=0.7cm]

\node[state, fill=black!5] (s_t) {$x_t$};
\node[state, fill=black!5, right=of s_t] (s_t_1) {$x_{t+1}$};
\node[state, fill=black!5, right=of s_t_1] (s_t_2) {$x_{t+2}$};
\node[state, fill=black!5, right=of s_t_2] (s_t_3) {$x_{t+3}$};
\node[state, fill=black!5, below=of s_t] (u_t) {$u_t$};
\node[state, fill=black!5, below=of s_t_1] (u_t_1) {$u_{t+1}$};
\node[state, fill=black!5, below=of s_t_2] (u_t_2) {$u_{t+2}$};
\node[state, fill=black!15, above=of s_t] (beta) {$\beta$};
\node[state, fill=black!5, above=of s_t_1] (obs_t_1) {$\hat{x}_{t+1}$};
\node[state, fill=black!5, above=of s_t_2] (obs_t_2) {$\hat{x}_{t+2}$};
\node[state, fill=black!5, above=of s_t_3] (obs_t_3) {$\hat{x}_{t+3}$};

\draw[-, line width=0.5mm] (beta) -- (s_t_1);
\draw[-, line width=0.5mm] (beta) -- (s_t_2);
\draw[-, line width=0.5mm] (beta) -- (s_t_3);
\draw[->, line width=0.1mm] (s_t) -- (s_t_1);
\draw[->, line width=0.1mm] (s_t_1) -- (s_t_2);
\draw[->, line width=0.1mm] (s_t_2) -- (s_t_3);
\draw[->, line width=0.1mm] (s_t_1) -- (obs_t_1);
\draw[->, line width=0.1mm] (s_t_2) -- (obs_t_2);
\draw[->, line width=0.1mm] (s_t_3) -- (obs_t_3);
\draw[->, line width=0.1mm] (u_t) -- (s_t_1);
\draw[->, line width=0.1mm] (u_t_1) -- (s_t_2);
\draw[->, line width=0.1mm] (u_t_2) -- (s_t_3);

\end{tikzpicture}
\caption{%
Graphical model for the proposed Gaussian process state-space model. Given a sequence of control inputs and the current state $x_t$, the distribution over future states is induced by $\beta$.
}
\label{fig:gpsm}
\end{figure}
}

%% file: sec_abstract.tex
\begin{abstract}
In order for robots to safely navigate in unseen scenarios using learning-based methods, it is important to accurately detect out-of-training-distribution (OoD) situations online.
Recently, Gaussian process state-space models (GPSSMs) have proven useful to discriminate unexpected observations by comparing them against probabilistic predictions.
However, the capability for the model to correctly distinguish between in- and out-of-training distribution observations hinges on the accuracy of these predictions, primarily affected by the class of functions the GPSSM kernel can represent.
%
%
In this paper, we propose (i) a novel approach to embed existing domain knowledge in the kernel and (ii) an OoD online runtime monitor, based on receding-horizon predictions.
Domain knowledge is provided in the form of a dataset, collected either in simulation or by using a nominal model.
%
Numerical results show that the informed kernel yields better regression quality with smaller datasets, as compared to standard kernel choices. We demonstrate the effectiveness of the OoD monitor on a real quadruped navigating an indoor setting, which reliably classifies previously unseen terrains.
\end{abstract}







%% file: sec_intro.tex
\section{Introduction}

As machine learning (ML) becomes increasingly integrated into autonomous robotic systems like service robotics \cite{brohan2023rt,brohan2023can} and self-driving cars \cite{nitsch2021out}, the need for reliable and safe ML models has never been more critical. Particularly in human-centric and safety-critical settings, even minor representation errors in ML models can have severe consequences \cite{salem2015towards}. 
Hence, it is crucial to identify the shortcomings of these models. One key issue is their potential unreliability when faced with data that differs from what they were trained on, commonly known as out-of-training-distribution (OoD).

This challenge has motivated the emerging research area of OoD detection \cite{yang2021generalized}. The field encompasses a variety of methodologies, ranging from invariant representation learning \cite{zhang2020learning,guo2023out} to causal learning for distribution shifts \cite{bagi2023generative}. While these techniques have been extensively studied in image classification \cite{yang2023bootstrap,de2023value}, their application to all levels of the full autonomy stack is still unfolding \cite{sharma2021sketching}.



%

In the field of robotics, real-time OoD detection is crucial for both real-time decision making and long-term reliability \cite{sinha2022system}, especially in uncertain environments.
Algorithms need to be uncertainty-aware and equipped with fallback strategies for OoD scenarios, either issuing warnings or adopting more conservative behaviors. However, research avenues in real-time OoD detection are rather sparse. In \cite{luo2022sample}, an OoD detector is trained on grasping datasets to have a provable low false negative rate. A similar idea is presented in \cite{farid2022failure}, where a failure predictor with guaranteed bounds is deployed on a real drone. However, these are not real-time, but rather sequential OoD monitors, since a full episode needs to be observed before OoD can be assessed. The same episodic scheme is followed in \cite{farid2022task}, where
OoD is measured using a task-specific performance cost. Our approach monitors OoD in real-time, which is crucial for decision-making in safety-critical situations, while remaining task-agnostic.

\begin{figure}[t!]
\centering
\includegraphics[width=0.8\columnwidth]{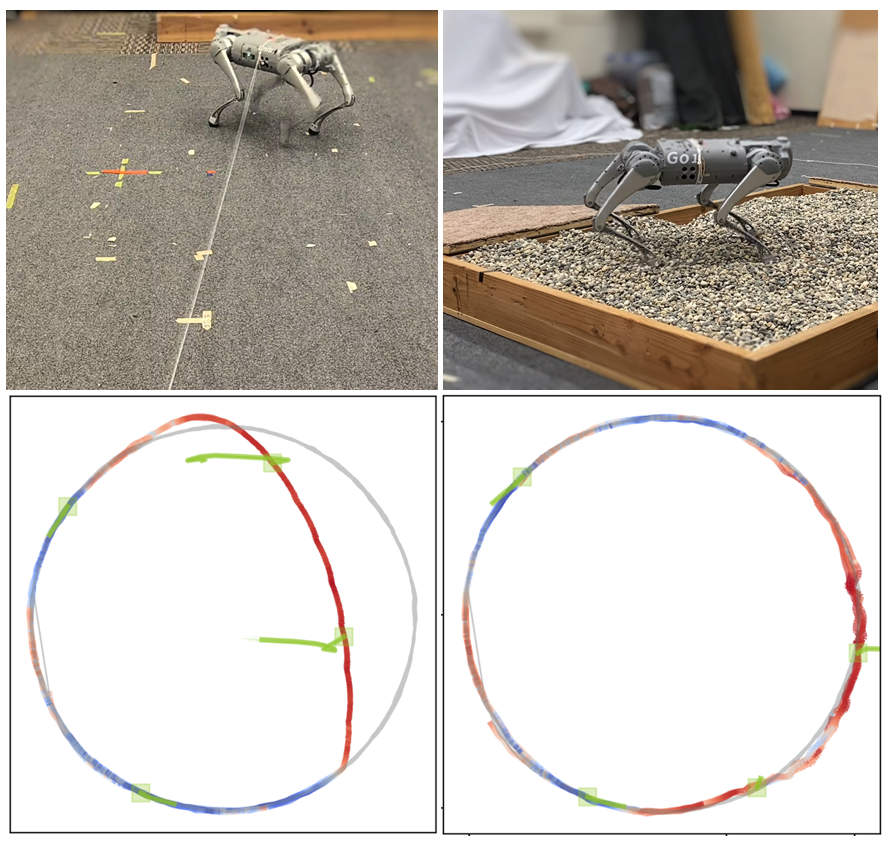}
\caption{A quadruped is exposed to different scenarios, unseen at training time, in order to assess the capability for out-of-training-distribution detection.}
\label{fig:eye_catcher}
\end{figure}

Real-time OoD runtime monitors offer significant advantages in the field of legged locomotion, where recognizing changes in environmental conditions is paramount to anticipate hazards.
Yet, deploying these monitors in highly dynamical systems, such as quadrupeds, poses major computational and design challenges.
%
%
In this work, we propose an OoD runtime monitor that tracks prediction errors in a receding horizon manner using Gaussian Process State-Space Models (GPSSMs) \cite{svensson2015nonlinear,salimbeni2017doubly,buisson2020actively,frigola2014variational,svensson2016computationally,eleftheriadis2017identification}. 
These models provide a probabilistic representation of the state-space dynamics and are well-suited for predicting future states and their uncertainties, making them ideal for safe navigation \cite{sun2021uncertain,zhu2020multi}.



Although GPSSMs have gained popularity for their ability to forecast state trajectories, they come with a few drawbacks. On one hand, the distribution of forecast states is often intractable and requires numerical approximations. On the other hand, they require large datasets for training, which is often impractical in real-world robotics where data is scarce \cite{eleftheriadis2017identification}.
%
To address the first issue, variational inference approaches have been proposed to also reduce the computational complexity \cite{eleftheriadis2017identification,frigola2014variational}, yet large datasets are still required.
One effective strategy to alleviate the second issue is to incorporate prior knowledge into the GPSSM's kernel function. Physics-aware models serve this purpose by using first-principles as regularizers \cite{wang2021physics}. Yet, these models often employ off-the-shelf kernels, overlooking the potential benefits of embedding existing knowledge in the kernel design. Recent work suggests that customizing the kernel for specific control problems can improve the model's performance \cite{marco2017design,sun2021uncertain}.
In this work, we tackle both issues by formulating the GPSSM as a Karhunen–Loève (KL) series expansion \cite{kanagawa2018gaussian,svensson2017flexible,svensson2016computationally,solin2020hilbert}, which poses two key advantages: (i) it enables efficient sampling schemes for state distribution forecasting, and (ii) it provides a flexible framework for designing kernels that integrate prior knowledge.

\emph{Contributions:} Our first contribution is a data-driven method for constructing GPSSM kernels that incorporates existing knowledge in the form of a queryable nominal dynamics model, such as a simulator or a first-principles representation. 
The key step is embedding the nominal model into the kernel via its Fourier representation. To this end, we minimize an autoencoder loss that reconstructs the nominal model.
The resulting GPSSM requires less data to achieve the same prediction accuracy as with standard kernels.



Our second contribution is an OoD runtime monitor that automatically detects scenarios unseen at training time. Our experiments on a real quadruped show that by leveraging the proposed GPSSM, the OoD runtime monitor can more accurately detect new terrains as compared to other GPSSMs that use standard non-informed kernels (e.g., Gaussian, or Matérn \cite{Rasmussen2006Gaussian}).
%
To the best of our knowledge, our work is the first to explicitly design a deployable online runtime monitor for OoD detection on a real quadrupedal robot.

The paper is structured as follows. In \cref{sec:prelim}, we explain the proposed GPSSM as a KL series expansion.  In \cref{sec:kernel}, we detail the proposed methodology to construct a kernel by embedding domain knowledge, given as a nominal dynamics model. In \cref{sec:oodmetric}, we explain how state forecasting can be used to detect OoD situations.
Finally, in \cref{sec:results}, we evaluate empirically the capabilities of the proposed model for OoD detection on a real quadruped, which navigates a room while being exposed to various changes in the environment, such as rocky terrain and external forces.

%% file: sec_prelim.tex
\section{Gaussian process state-space model using Mercer kernels}
\label{sec:prelim}
\noindent Let $x_t \in \mathcal{X} \subseteq \R^D$ be the system state and $u_t \in \mathcal{U} \subseteq \R^{D_u}$ the control input at time $t$. The transition to the next state $x_{t+1}$ follows the system dynamics $f_\textrm{true} : \mathcal{X} \times \mathcal{U} \rightarrow \mathcal{X}$. 
%
%
Whereas the true dynamics are unknown, they can be represented by a dynamics model $f(\cdot) = [f_1(\cdot),\ldots,f_D(\cdot)]^\top$, which is described as a Gaussian process state-space model (GPSSM) \cite{frigola2014variational}
\begin{gather}
\begin{aligned}
f_d(\cdot) & \sim \mathcal{GP}(m_d(\cdot),k_d(\cdot,\cdot)),\; d = 1,\ldots,D \\
x_{t+1} & \sim \mathcal{N}(f(x_t,u_t),Q) \\ 
\hat{x}_{t+1} & \sim p(\hat{x}_{t+1} | x_{t+1}),
\label{eq:mod}
\end{aligned}
\end{gather}
where each component $f_d : \mathcal{X} \times \mathcal{U} \rightarrow \R$ follows an independent Gaussian process (GP) with kernel $k_d : \mathcal{X} \times \mathcal{X} \rightarrow \R$ and mean function $m_d : \mathcal{X} \rightarrow \R$, and $Q = \text{diag}([\sigma_1^2,\ldots,\sigma^2_D])$. For simplicity, we assume a Gaussian observation model\footnote{While integrating non-Gaussian observation models is possible, this is known to be challenging \cite{eleftheriadis2017identification,frigola2014variational} and out of the scope of this paper.} $\hat{x}_{t+1} \sim \mathcal{N}(x_{t+1},\sigma^2_\textrm{n}I)$.

In the following, we focus on modeling a single component of the dynamics $f_d$ without loss of generality. For ease of presentation, we denote $z_t = [x_t^\top,u_t^\top]^\top$ and $\mathcal{Z} = \mathcal{X} \times \mathcal{U}$. We drop the subindex $d$ in both $f_d$ and $\sigma_d^{2}$ and define the $d$ component of a state observation as $X_t = [\hat{x}_t]_d$.

\subsection{Karhunen–Loève expansion}
\label{ssec:KLexp}
The Gaussian process (GP) representing the dynamics $f$ can be defined in weight space as \cite[Sec. 2.1]{Rasmussen2006Gaussian}
\begin{align}
f(\zt) & = \sum_{j=1}^M \beta_{j} \phi_{j}(\zt),
\label{eq:KLmodel}
\end{align}
with $M > 0$ independently distributed random weights $\beta_j \sim \mathcal{N}(m_j,v_j)$, mean $m_j \in \R$ and variance $v_j > 0$, and deterministic features $\phi_j : \mathcal{Z} \rightarrow \R$. This formulation is known as the \emph{Karhunen–Loève (KL) series  expansion} of a GP \cite{kanagawa2018gaussian}.
%
%
The covariance function associated with \cref{eq:KLmodel} is given as \cite{minka2000deriving}
\begin{align}
k(\zt,\ztp) = \sum_{j=1}^M v_j \phi_{j}(\zt) \phi_{j}(\ztp),
\label{eq:kermercer}
\end{align}
which is a (positive definite) \emph{Mercer kernel} \cite{kanagawa2018gaussian}.
%
%
Informally, the KL expansion \cref{eq:KLmodel} is equivalent to a standard GP formulation $\mathcal{GP}(m(z_t),k(\zt,\ztp))$ with a positive definite kernel $k$ that admits a series representation \cref{eq:kermercer}.

\subsection{Predictive posterior}
\label{ssec:predpost}
We are interested in the above formulation to predict the dynamics at new locations $f(z_*)$. 
Let us assume access to an approximate description of the dynamics through a nominal model $\fnom : \mathcal{Z} \rightarrow \R$ that captures the general behavior of the system, e.g., a simulator. We use $\fnom$ to acquire a dataset of $T$ state-control-state tuples $\mathcal{D} = \{\hat{z}_t,X_{t+1}\}_{t=0}^{T-1} \sim \nu(z_t)$, where the measure $\nu(z_t)$ is given implicitly as a distribution over dynamically feasible trajectories in $\mathcal{Z}$, and the observations are $X_{t+1} = \fnom(\hat{z}_t)$.
%
%
%
%
The predictive posterior is solely determined by the posterior over the weights $p(\beta | \mathcal{D}) = \mathcal{N}(\mu,\Sigma)$, which is given analytically \cite[Sec. 9.3]{deisenroth2020mathematics} by
\begin{gather}
\begin{aligned}
\mu = & \Sigma (V^{-1} m  + \sigma^{-2}\Phi_Z X)  \\
\Sigma = & (V^{-1} + \sigma^{-2}\Phi_Z^{ }\Phi_Z^\top)^{-1},
\label{eq:post}
\end{aligned}
\end{gather}
with feature matrix $\Phi_Z = [\Phi(\hat{z}_0),\ldots,\Phi(\hat{z}_{T-1})]$, feature vector
$\Phi(\hat{z}_t) = [ \phi_1(\hat{z}_t),\ldots,\phi_M(\hat{z}_t) ]^\top$, observations
$X=[X_1,\ldots,X_T]^\top$, and prior parameters
$m = [m_1,\ldots,m_M]^\top$ and $V = \text{diag}([v_1,\ldots,v_M])$. The predictive posterior over $f(z_*)$ at a new location $z_*$ is
\begin{align}
f(z_*) | \mathcal{D} & \sim \mathcal{N}\left(\mu^\top \Phi_*, \Phi^\top_*\Sigma \Phi_* \right),\;\; \Phi_* = \Phi(z_*).
\label{eq:predpost}
\end{align}


\subsection{Computational complexity and model expressiveness}
In general, a larger number of features $M$ in \cref{eq:kermercer} produces a more expressive model \cite{marco2017design}. Specifically, if $M \rightarrow \infty$, the series in \cref{eq:kermercer} converges to a closed-form expression \cite{minka2000deriving,marco2017design} for particular feature choices (e.g., Gaussian-shaped features give rise to the squared exponential kernel \cite[Sec. 4.2.1]{Rasmussen2006Gaussian}). Since such closed-form kernels entangle infinitely many features, they yield more expressive models than their truncated counterpart. 
However, the GP posterior with such kernels suffers from $O(T^3)$ complexity. On the contrary, the truncated series formulation \cref{eq:KLmodel} poses two main benefits: (i) it has cost $O(M^3T)$, i.e., linear with the number of datapoints\footnote{In practice, we keep $M$ as large as computational resources allow.} and (ii) it yields flexibility in constructing Mercer kernels that integrate existing domain knowledge. We discuss this idea next.

%% file: sec_kernelconstruction.tex
\section{Constructing a data-driven domain informed kernel}
\label{sec:kernel}
\noindent In this section we propose a methodology to integrate existing domain knowledge into a Mercer kernel \cref{eq:kermercer}. 

\subsection{Integrating domain knowledge via Fourier features}
\label{ssec:domknow}
The nominal model $\fnom$ introduced in \cref{ssec:predpost} holds prior information about the dynamics. Our approach is integrating such model
into the GPSSMs as the prior mean function \cite{frigola2013bayesian} through the mean coefficients of the KL expansion \cref{eq:KLmodel}, i.e.,
\begin{align}
\mathbb{E}\left[ \sum_{j=1}^M\beta_j \phi_j(z_t) \right] = \sum_{j=1}^M m_j \phi_j(z_t) = \fnom(z_t).
\label{eq:series}
\end{align}
While there exist many viable function families to represent $\phi_j(z_t)$, we choose a Fourier series expansion \cite{zygmund2002trigonometric} due to its well-studied connection with the frequency domain \cite{Rasmussen2006Gaussian}.

Let $\fnom$ be square integrable with respect to $\nu$.
Then, its truncated multivariate Fourier series expansion can be written as a neural network with one hidden layer \cite{ngom2021fourier} and a cosine activation function
\begin{align}
\fnom(z_t) = \sum_{j=1}^M S(\omega_j) \cos(\omega_j^\top z_t + \varphi(\omega_j)),
\label{eq:fourierseries}
\end{align}
where the frequencies $\omega_j \in \Omega \subseteq \R^{D + D_u}$ are spaced in a regular grid $\omega_j = \tilde{\omega}[j_1,\ldots,j_{D_z}]^\top,\; j_i \in \mathbb{Z}$, and $\tilde{\omega} > 0$ is a base frequency. The modulus $S(\cdot) = | \mathcal{F}[\fnom](\cdot) |$ is known as \emph{spectral density} \cite{bracewell1986fourier} and $\varphi(\cdot) = \angle \mathcal{F}[\fnom](\cdot)$ is the \emph{system phase}. 
%
They both depend on the Fourier series coefficients $\mathcal{F}[\fnom](\omega_j)$, defined through the multivariate Fourier transform \cite{bracewell1986fourier}
\begin{align}
\mathcal{F}[\fnom](\omega_j) = & \int_{\mathcal{Z}} \fnom(z_t)e^{-i\omega_j^\top z_t}\text{d}\nu(z_t).
\label{eq:fouriercoeff}
\end{align}


This approach poses two main computational caveats. First, having the frequencies in \cref{eq:fourierseries} regularly distributed is problematic because (i) the grid size $M$ increases exponentially with the dimensionality, and (ii), an uncareful choice of grid size may result in suboptimal placement of frequencies in regions where the influence of $S(\omega_j)$ and $\varphi(\omega_j)$ is negligible\footnote{The Riemann–Lebesgue lemma \cite{katznelson2004introduction} establishes that $S(\omega)$ vanishes as $|\omega| \rightarrow \infty$.}.
Second, the integrand in \cref{eq:fouriercoeff} is only observable at sparse locations, sampled from $\nu(z_t)$ (see \cref{ssec:predpost}). In the following, we propose a numerical approach to mitigate both issues. We denote $S_j = S(\omega_j)$ and $\varphi_j = \varphi(\omega_j)$.
%

To alleviate the first issue, we propose an irregular grid that increases resolution in areas where $S_j$ is large. Specifically, we follow \cite{ngom2021fourier}, where the optimal parameters $\kappa^{*} = \{S_j^{*}, \varphi_j^{*}, \omega_j^{*}\}_{j=1}^M$ are obtained by minimizing the loss
\begin{align}
\min_\kappa \dfrac{1}{T} \sum_{t=1}^{T-1}|| X_{t+1} - \sum_{j=1}^M S_j \cos(\omega_j^\top \hat{z}_t + \varphi_j)  ||_2^2 + \lambda_\omega||W||_2^2,
\label{eq:recloss}
\end{align}
where $W = [\omega_1,\ldots,\omega_M ]$ regularizes the frequencies. With this approach, the decoder network \cref{eq:fourierseries} shall place the learned frequencies $\omega^{*}_j$ in a non-uniform grid with higher resolution in areas where $S^{*}_j$ and $\varphi^{*}_j$ are most influential. The reconstruction loss is evaluated on the dataset $\mathcal{D}$ (see \cref{ssec:predpost}).

We tackle the second issue by numerically approximating \cref{eq:fouriercoeff} as a quadrature $\sum_{t=1}^{T-1} X_{t+1}e^{-i\omega_j^\top \hat{z}_t} \eta_t$, which can be seen as a one-layer encoder network, where the integration steps $\eta_t > 0$ are jointly learned with $\kappa$.

\begin{figure}[t]
\centering
\includegraphics[width=1\columnwidth]{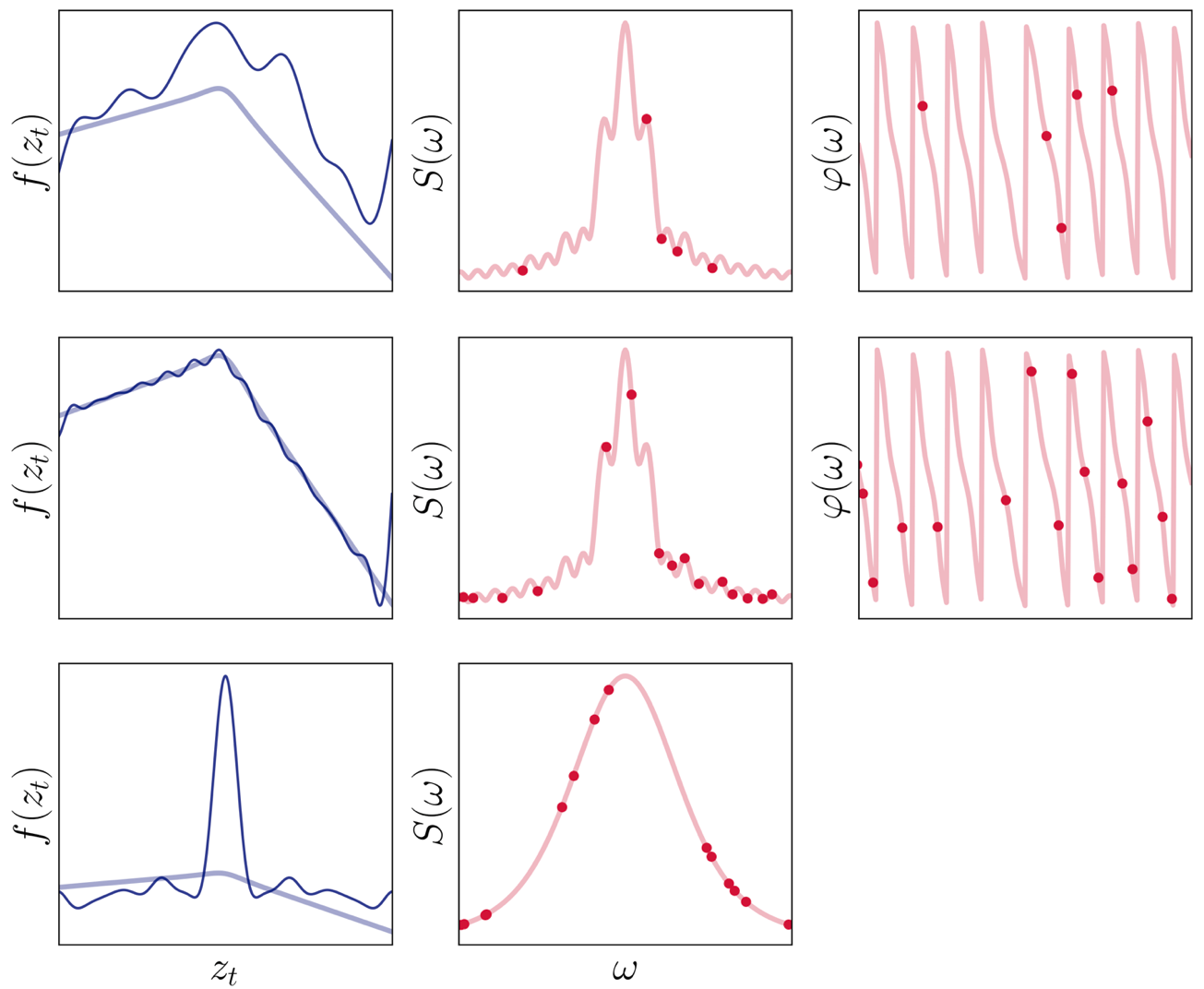}
\caption{Reconstruction of the ``elbow'' dynamical system. Top row, first column: True dynamical system (light blue line) and reconstructed dynamical system (dark blue line) with $M=5$ features. Second column: Spectral density $S(\omega)$ of the true dynamical system. Third column: System phase $\varphi(\omega)$ of the true dynamical system. The red dots indicate the optimal values $S^{*}_j$, $\varphi^{*}_j$ found after optimizing \cref{eq:recloss}. Middle row: Analogous to top row with $M=20$. Bottom row: Reconstruction using the spectral density of a Matérn 3/2 kernel.}
\label{fig:elbow}
\end{figure}

The sought mean coefficients and features in \cref{eq:series} are
\begin{gather}
\begin{aligned}
m_j = S^{*}_j,\;\;\;
\phi_j(z_t) & = \cos((\omega^{*}_j)^\top z_t + \varphi_j^{*}).
\label{eq:features}
\end{aligned}
\end{gather}
\noindent
The reconstruction quality is affected by the size $M$ of the irregular grid. We illustrate in \cref{fig:elbow} the impact of different grid sizes using the following toy example:
\begin{example} 
Let us consider the scalar ``elbow'' dynamical system $f(z_t) = 0.8 + (z_t + 0.2)(1 - 0.5(1 + \text{exp}(-2z_t))^{-1})$ \cite{ialongo2019overcoming},
\label{ex:elbow}
which we wish to reconstruct using the aforementioned autoencoder.
To this end, a dataset of size $T=50$ is collected from this dynamical system within the interval $z_t \in [-10,10]$.
The first two rows show the reconstructed $\fnom$ after optimizing the autoencoder. The reconstruction improves as we increase the number of features from $M=5$ to $M=20$ for a small number of optimization steps. We also depict the true spectral density $S(\omega)$ and system phase $\varphi(\omega)$ that uniquely characterize this dynamical system. After optimization, the optimal frequencies $\omega_j^{*}$ are found at locations where both $S^{*}_j$ and $\varphi^{*}_j$ contribute the most to the reconstruction.
\end{example}
The proposed formulation allows us to embed information about $\fnom$ in the kernel \cref{eq:kermercer} through the dataset $\mathcal{D}_\textrm{n}$, which is discussed next.

\subsection{Non-stationary domain-informed Mercer kernel}
\label{ssec:kerdom}
\noindent Given the informed features and mean coefficients \cref{eq:features}, the proposed kernel follows from \cref{eq:kermercer}
\begin{align}
k(z_t,z_t^\prime) = \sum_{j=1}^M v_j \cos((\omega^{*}_j)^\top z_t + \varphi_j^{*}) \cos((\omega^{*}_j) z^\prime_t + \varphi_j^{*}).
\label{eq:kerours}
\end{align}
A key difference between this kernel and commonly used kernels (e.g., Gaussian or Matérn \cite[Sec. 4.2]{Rasmussen2006Gaussian}) or similar kernels with harmonic-based series representations \cite{solin2020hilbert,hensman2017variational,rahimi2007random}, is that it is generally non-stationary, i.e., $k(z_t,z_t^\prime) \neq k(z_t-z_t^\prime)$.
We find stationarity to be a restricting assumption, as it implies that the statistical properties of $\fnom$ do not change over the input space, which is not necessarily true for arbitrary dynamical systems. The non-stationarity in \cref{eq:kerours} adds flexibility to the model without necessarily overfitting to a specific function class.
%
We illustrate the shortcomings of stationarity in \cref{fig:elbow} (third row), where we attempt to reconstruct $\fnom$ by using a Matérn kernel, which is an isotropic stationary kernel. To this end, we first transform \cref{eq:kerours} into a stationary kernel by fixing $v_j = \sqrt{2}$ and $\varphi_j \sim \textrm{U}(0,2\pi)$, as in \cite{rahimi2007random}. Then, the optimal frequencies $\omega_j^{*}$ are obtained by solving \cref{eq:recloss}, and setting $S_j^{*} = S_\text{Matérn}(\omega_j^{*})$, where $S_\text{Matérn}(\cdot)$ is the spectral density of a Matérn kernel, given in \cite[Sec. 4.2]{Rasmussen2006Gaussian}. As shown, a non-informed stationary kernel alone is unable to capture the properties of the true dynamical system.


The proposed kernel enables embedding prior information into the proposed GPSSM \cref{eq:mod}, which shall influence the long-term state predictions. In the following, we discuss how to use informed receding-horizon predictions to design an out-of-training-distribution runtime monitor.

%% file: sec_problem.tex
\section{OoD runtime monitor with GPSSM}
\label{sec:oodmetric}
We are interested in quantifying how likely it is that the robot navigates a scenario that is out-of-training-distribution (OoD). In the following, we propose a runtime monitor that compares the distribution of predicted states with the 
posterior after conditioning on collected observations.

\subsection{Prediction-based OoD detection}



Let $u_{t:H-1} = \{u_t,\ldots,u_{t+H-1}\}$ be a control sequence that rolls out the dynamical model forward in time $H$ steps, departing from the current state $x_t$. The resulting latent state sequence $x_{t+1:H}$, assumed to be Markovian, is distributed as
\begin{align}
q(x_{t+1:H} | x_t, u_{t:H-1}) = \prod_{h=0}^{H-1}p(x_{t+h+1} | x_{t+h}, u_{t+h}).
\label{eq:probstates}
\end{align}
Such distribution determines the future model predictions from the current state $x_t$. We let the system evolve for $H$ time steps under $u_{t:H-1}$, and collect a set of observations $\hat{x}_{t+1:H}$, which induce a posterior distribution over the states $p(x_{t+1:H} | \hat{x}_{t+1:H}, x_t, u_{t:H-1})$. Our goal is to use the Kullback–Leibler divergence $D_\text{KL}(q || p) > 0$ as a proxy to detect OoD. The intuition behind this approach is as follows: In-distribution scenarios should be accurately predicted, resulting in a small divergence, while OoD scenarios should report a large divergence. Furthermore, it is well known that a low divergence $D_\text{KL}(q || p)$ yields a large evidence lower bound (ELBO) while the (unknown) log-evidence remains constant, i.e., $D_\text{KL}(q || p) + \mathbb{E}_{q}\left[ \log p(\hat{x}_{t+1:H} | x_{t+1:H}) \right] = \log p(\hat{x}_{t+1:H})$.
Therefore, 
we define the OoD loss as the negative evidence lower bound
%
%
\begin{align}
\mathcal{L}_\text{OoD} = -\mathbb{E}_{q}\left[ \log p(\hat{x}_{t+1:H} | x_{t+1:H}) \right].
\label{eq:elboroll}
\end{align}

\noindent
The distribution over the predicted state sequence \cref{eq:probstates} is non-Gaussian \cite{eleftheriadis2017identification}, which renders the expectation in \cref{eq:elboroll} analytically intractable. In practice, we approximate it using a Monte Carlo sum \cite{svensson2016computationally,svensson2017flexible}. Since the likelihood \cref{eq:mod} factorizes, the approximated runtime monitor is
\begin{align}
\mathcal{L}_\text{OoD} \approx \frac{1}{R} \sum_{r=1}^R\sum_{h=1}^{H}\dfrac{1}{2\sigma_\text{n}^2} ||(\hat{x}_{t+h} - \tilde{x}^{(r)}_{t+h}||_2^2 + \text{const.},
\label{eq:oodlosstot}
\end{align}
with rollouts $\tilde{x}^{(r)}_{t+1:H} \sim q(x_{t+1:H} | x_t, u_{t:H-1})$.
We detail the sampling approach next.


\subsection{Sampling scheme}
\label{ssec:sampling}
Given a control sequence $u_{t:H-1}$, our goal is sampling a future state sequence from \cref{eq:probstates}. While this is generally challenging, a key advantage of \cref{eq:mod} over vanilla GPSSM is that $\beta_d,\;d=\{1,\ldots,D\}$ constitutes sufficient statistics for predicting future states (see \cref{fig:gpsm}). Specifically, a posterior sample $\tilde{\beta}_d \sim \mathcal{N}(\mu_d,\Sigma_d)$ constitutes a deterministic instance of the unknown dynamics $\tilde{f}_d(x_t,u_t) = \tilde{\beta}^\top_d \Phi_d(x_t, u_t),\; d=1,\ldots,D$. In practice, we use the reparametrization trick $\tilde{\beta}_d^{(r)} = \mu_d + L_d \tilde{\eta}^{(r)},\;r=\{1,\ldots,R\}$ to pre-sample and store $R$ model instantiations at a fixed cost, where $L_d$ is obtained through the Cholesky decomposition $\Sigma_d = L_d L_d^\top$ and $\tilde{\eta}^{(r)} \sim \mathcal{N}(0,1)$. Under the Gaussian observation model \cref{eq:mod}, the next state is given deterministically as
\begin{align}
\tilde{x}_{t+1}^{(r)}(x_t,u_t) = 
\begin{bmatrix}
(\tilde{\beta}_1^{(r)})^\top \Phi_1(x_t, u_t) \\ 
\vdots \\
(\tilde{\beta}_D^{(r)})^\top \Phi_D(x_t, u_t)
\end{bmatrix}
+
\tilde{\gamma}^{(r)},
\label{eq:nextstate}
\end{align}
with $\tilde{\gamma}^{(r)} \sim \mathcal{N}(0,Q+\sigma^2_\textrm{n}I)$. Hence, a sequence of $H$ states is obtained by recursively calling \cref{eq:nextstate}. The overall runtime cost of sampling $R$ trajectories is $O(HDR)$ per time step. 


\graphmodel{t}

%% file: sec_results.tex
\section{Results}
\label{sec:results}
In this section, we present two sets of results. First, we assess the capability of the proposed method to detect out-of-training-distribution (OoD) situations with a real quadruped. We compare the performance of our GPSSM with another model that uses standard non-informed kernels. Second, we present a brief ablation study where we show that our domain-informed kernel is indeed more data efficient than existing GPSSMs that use standard kernels.

\emph{Parameter choices:} The prior variance $v_j$ over the weights \cref{eq:mod} is a design choice that affects the quality of the regression. In practice, one can choose any $v_j > 0$ set of eigenvalues, e.g., constant \cite{hensman2017variational}. Herein, we follow \cite{solin2020hilbert} and set $v_j = \xi m_j$, with $\xi > 0$, i.e.,
proportional to the spectral density value (see \cref{eq:features}).
%
Intuitively, this choice assigns high confidence ($v_j \approx 0$) to weakly weighted features (i.e., $m_j \approx 0$), which causes them to have low impact in the model, while strongly weighted features remain uncertain, yet flexible and adaptable to the posterior.
We utilize GPyTorch \cite{gardner2018gpytorch} to train the non-informed GPSSMs that employ standard kernels, conducting hyperparameter optimization for the same number of epochs as we do for the proposed model.

%

%

\subsection{OoD runtime monitor in quadrupedal locomotion}
In order to assess the efficacy of the proposed OoD runtime monitor, we conduct experiments using the quadrupedal robot Go1 \cite{unitree23} in an indoor environment. Our goal is to detect various changes in the environment as the robot traverses a circular path. To this end, we first gather motion data as the robot walks in circles, without any environmental perturbations. Subsequently, we train the domain-informed GPSSM (\cref{sec:prelim,sec:kernel}) using the collected data. Finally, we deploy the OoD runtime monitor described in \cref{sec:oodmetric}, which leverages the trained model and detects environment changes. The goal is for the OoD runtime monitor to correctly identify the portions of the path affected by a variety of environment changes, which we detail next.



\emph{Rope pulling:} A rope is attached to the robot's body while the other end of the rope is anchored, inducing a pulling force that deviates the robot from its intended circular path for a certain portion of the trajectory (see \cref{fig:eye_catcher}).


\emph{Rocky terrain:} The robot is required to navigate through rocky terrain for a section of the trajectory.

\emph{Poking:} The robot's body is gently prodded with a pole at various stages during its movement.

In the following, we describe our experimental setup, design choices, and discuss the results.


\subsubsection{Design choices}
We use the same state and control input representation as the ones for unicycle dynamics \cite{molnar2023safety}. The state $x_t = [\mathrm{x}_t, \mathrm{y}_t, \alpha_t]^\top$ concatenates the X-Y robot position in the room and the heading angle $\alpha_t \in [-\pi,\pi)$, i.e., the robot's orientation in the X-Y plane. The control input $u_t = [\mathrm{v}_t,\dot{\alpha}_t]^\top$ gathers the desired forward velocity $\mathrm{v}_t$ and the desired rotation speed $\dot{\alpha}_t$.
Due to several real-world effects, the unicycle dynamics \cite{molnar2023safety} are affected by several nonlinearities that are difficult to model, for example, communication delays and sensor noise. Also, inherent dynamics in the manufacturer's walking algorithms forbid from accurately tracking the desired speeds. The proposed GPSSM \cref{eq:mod} absorbs such nonlinearities. We learn each component $f_d$ using $M=1500$ features and $\sigma_d = 0.01,\; d = \{1,2,3\}$.

\emph{Autoencoder training:} As detailed in \cref{ssec:domknow}, the autoencoder reconstructs a given nominal dynamics model $\fnom$. In this setting, the nominal model is implicitly represented through a dataset $\mathcal{D}$ of state-control-state tuples, collected directly on the real system. Specifically, we collect 10 minutes of data at 10 Hz of the robot walking in circles on flat ground and use this dataset to train the autoencoder with $\lambda_\omega = 0.1$. We train for 100,000 epochs with a cosine decay learning rate. The frequencies $\omega_j$ are initialized by sampling from a uniform distribution within the domain $[-3,3]^5$.

\emph{Sampling parameters:} We follow the sampling scheme proposed in \cref{ssec:sampling} to compute \cref{eq:oodlosstot}. At each time step, we sample $R=20$ system rollouts with a time horizon of $H=30$ timesteps (i.e., 3 seconds lookahead).




\subsubsection{Experimental setup}
The position and orientation of the robot are estimated using a VICON motion tracker. The desired speed commands are computed online using a proportional controller that steers the robot toward a set of 
%
20 waypoints distributed along a circle measuring 3 meters in diameter. The desired forward velocity is given as
$\mathrm{v}^\text{des}_t = 0.1 + K_p || (\mathrm{x}_\text{next},\mathrm{y}_\text{next}) - (\mathrm{x}_t, \mathrm{y}_t) ||_2$, where $(\mathrm{x}_\text{next},\mathrm{y}_\text{next})$ is the position of the next waypoint. The desired angular velocity is computed as  $\dot{\alpha}^\text{des}_t = 0.2 (\theta_\text{next} - \theta_t)$. This controller is fixed across all environments.
The OoD loss is normalized to be $\mathcal{L}_\text{OoD} \in [0,1]$ across all environments. We send desired velocity commands to the robot's on-board computer at 100 Hz via wireless communication. Our implementation uses ROS in Python and C++ on a standard commercial laptop.

\begin{figure}[t!]
\centering
\includegraphics[width=0.9\columnwidth]{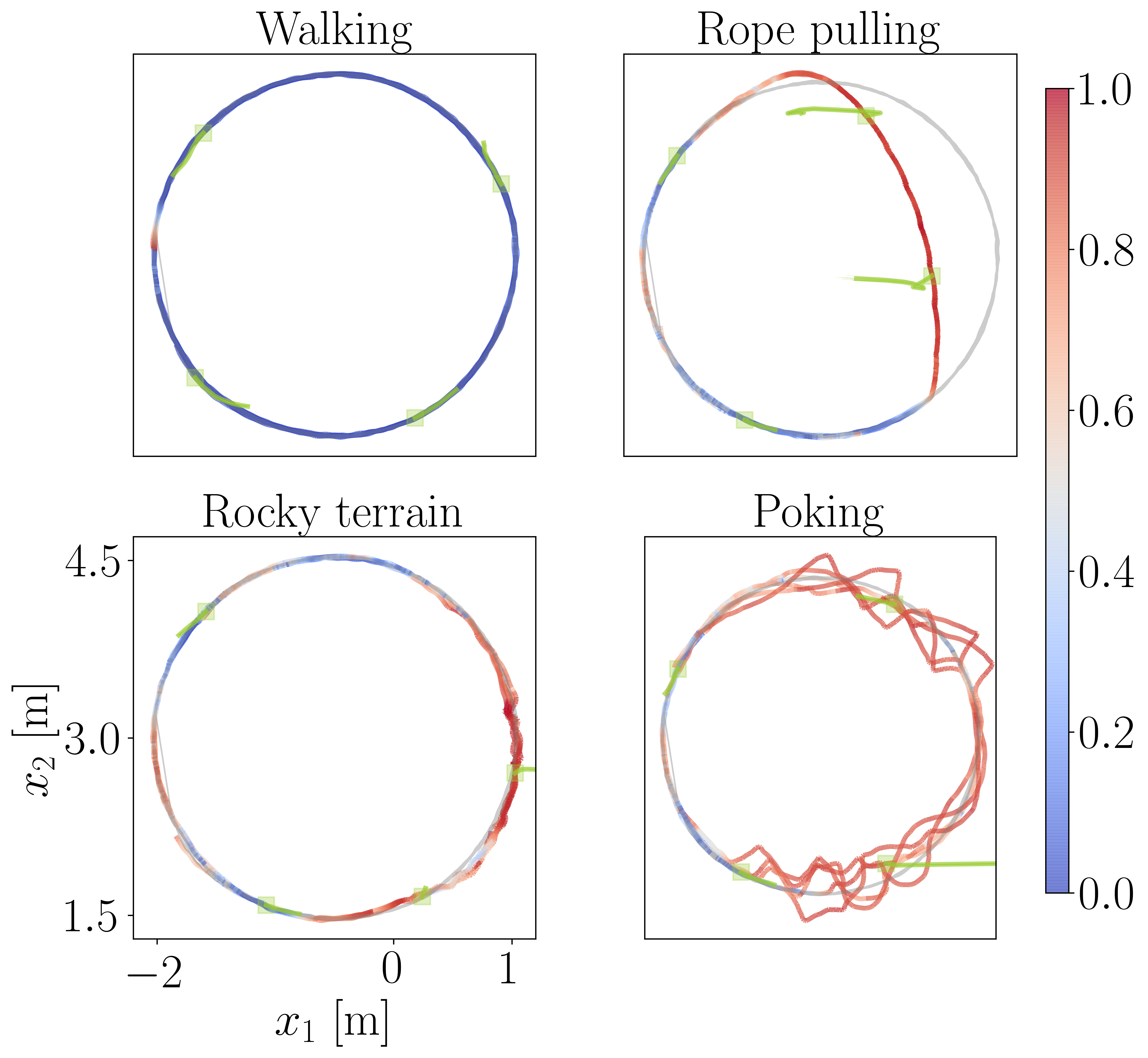}
\caption{A quadrupedal robot follows an anti-clockwise circular motion, while being exposed to three changes in the environment. Top-left: The robot walks in flat terrain without environment changes. Top-right: The robot is tied to a rope that creates a pulling force. Bottom-left: The robot walks on rocky terrain. Bottom-right: the robot is poked in random directions as it walks. The colormap transitions from blue (low OoD loss values) to red (high OoD loss values). Future state predictions are overlaid in light green at four different parts of the trajectory.}
\label{fig:oodquadruped}
\end{figure}

\subsubsection{Results discussion}
In \cref{fig:oodquadruped}, we show that $\mathcal{L}_\text{OoD}$ keeps low values when the robot passes through parts of the trajectory that were seen at training time. On the contrary, $\mathcal{L}_\text{OoD}$ reaches high values when the robot is exposed to changes in the environment not captured at training time. In the rope pulling case, we see that the sampled state trajectories deviate significantly from the trajectory when the rope is tense, but remain in-distribution when it is loose. We see similar situations in the other two settings.

\begin{table}[b!]
\centering
\caption{Percentage of time steps in which $\mathcal{L}_\text{OoD} > 0.5$ across all trajectories for each environment.}
\begin{tabular}{c|cccc}
\toprule
& \textbf{Walking} & \textbf{Rope} & \textbf{Rocky} & \textbf{Poking} \\
\midrule
\textbf{Ours} & \textbf{1.8\%} & \textbf{66.7\%} & \textbf{64.4\%} & \textbf{79.0\%} \\
GPSSM & 87\% & 92.5\% & 98.7\% & 97.3\% \\
\bottomrule
\end{tabular}
\label{tab:perctime}
\end{table}

These results indicate that our model consistently remains in-distribution when the environment is significantly similar to the one it was exposed to at training time. In \cref{tab:perctime} we show the percentage of time steps in which the robot was out of distribution. To determine this, we normalize the data offline and classify OoD situations with $\mathcal{L}_\text{OoD} > 0.5$. We compare these results with a Gaussian process state space model that uses a standard Matérn 3/2 kernel, trained with the same data from circular motions. As shown in \cref{tab:perctime}, this model's predictions remain in distribution less often. In addition, the model has difficulties predicting reliably the data it was trained on (i.e., first column). This indicates that for such a small dataset, informing the kernel is beneficial to yield more accurate predictions.
We found the number of features $M=1500$ to be a key factor in obtaining accurate state predictions. While a larger $M$ did not yield better results, smaller $M$ affected the quality of the long-term predictions, which made the OoD runtime monitor unreliable.

\subsection{Data efficiency assessment}
It is desirable to analyze the extent to which the domain-informed kernel increases data efficiency. To assess this, we collect 10 minutes of data in the same setting described above, but moving the robot through a set of randomly chosen waypoints across the room. We leave 10\% of the dataset for testing and use the remaining portion as a training set. Then, we slice the training dataset in batches of increasing size (25\%, 50\%, 75\%) and train the kernel of the informed GPSSM with each of them. As a performance metric we use the root mean squared error (RMSE) between the predictive mean of the GPSSM and the actual observations to assess the quality of the regression fit. We compare the performance of our model with a standard GPSSM that uses (i) a squared exponential kernel (SE) and (ii) a Matérn kernel. In \cref{fig:kernelrec}, we see that our model systematically achieves a lower RMSE, which indicates that constructing the kernel by integrating existing data does increase data efficiency.

\begin{table}[t!]
\centering
\caption{Data efficiency analysis.}
\begin{tabular}{c|c|c|c|c}
\toprule
 & \textbf{25\%} & \textbf{50\%} & \textbf{75\%} & \textbf{100\%} \\
\midrule
\multicolumn{1}{c|}{SE} & -3.68 (0.09) & -3.72 (0.03) & -3.75 (0.01) & -3.47 (0.20) \\
\multicolumn{1}{c|}{Matern} & -3.74 (0.03) & -3.71 (0.05) & -3.69 (0.08) & -3.58 (0.17) \\
\multicolumn{1}{c|}{\textbf{Ours}} & \textbf{-4.35 (0.04)} & \textbf{-4.42 (0.02)} & \textbf{-4.49 (0.04)} & \textbf{-4.54 (0.03)} \\
\bottomrule
\end{tabular}
\label{fig:kernelrec}
\end{table}


%

%% file: sec_conclusions.tex
\section{Conclusion}
In this paper, we have proposed (i) a general methodology to embed domain knowledge, given as a nominal model, into the kernel of a Gaussian process state-space model (GPSSM), and (ii) a runtime monitor for online out-of-training-distribution (OoD) detection using GPSSMs.

The resulting non-stationary kernel encapsulates domain knowledge using Fourier features, which are learned by minimizing an autoencoder loss using data from an implicit nominal model. Our results indicate that such a domain-informed kernel needs less training data to achieve the same regression quality as standard non-informed stationary kernels. In addition, we show that the proposed OoD runtime monitor, validated on a quadruped robot navigating in an indoor setting, can reliably detect previously unseen scenarios. 

Whereas this work focuses on detecting OoD situations online, we plan to integrate the OoD runtime monitor with sampling-based model predictive control methods \cite{williams2016aggressive} for online decision-making, to enable safe navigation.

\section*{Acknowledgment}
The authors of this paper express their gratitude to Jason J. Choi and Chams E. Mballo for insightful discussions, and to Prof. Koushil Sreenath for his support in providing computational resources.